\documentclass[10pt,twocolumn,letterpaper]{article}

\usepackage[pagenumbers]{cvpr} % To force page numbers, e.g. for an arXiv version

\usepackage{graphicx}
\usepackage{amsmath}
\usepackage{amssymb}
\usepackage{booktabs}
\usepackage[pagebackref,breaklinks,colorlinks]{hyperref}

\usepackage{latexsym}
\usepackage{booktabs}
\usepackage{algorithm}
\usepackage{algorithmic}
\usepackage[switch]{lineno}

\usepackage{multirow}
\usepackage{makecell}
\usepackage{pifont}
\usepackage{colortbl}
\usepackage{xcolor}
\usepackage{subcaption}
\usepackage{diagbox}
\usepackage{bm}
\usepackage{epigraph}

\definecolor{demphcolor1}{gray}{.6}
\definecolor{sherrybrown}{RGB}{227,207,87}
\definecolor{lightred}{RGB}{241,140,142}
\definecolor{deepgreen}{RGB}{127, 161, 104}
\definecolor{lightblue}{RGB}{122,46,246}

\newcommand\blfootnote[1]{%
  \begingroup
  \renewcommand\thefootnote{}\footnote{#1}%
  \addtocounter{footnote}{-1}%
  \endgroup
}

\usepackage[capitalize]{cleveref}
\crefname{section}{Sec.}{Secs.}
\Crefname{section}{Section}{Sections}
\Crefname{table}{Table}{Tables}
\crefname{table}{Tab.}{Tabs.}

\def\cvprPaperID{*****} % *** Enter the CVPR Paper ID here
\def\confName{CVPR}
\def\confYear{2022}

\begin{document}

% \title{\includegraphics[width=0.4cm]{figure/Vermouth_icon.pdf} Bridging Generative and Discriminative Models for Unified Visual Perception with Diffusion Priors}
\title{Foodfusion: A Novel Approach for Food Image Composition \\ via Diffusion Models}

% \author{Chaohua Shi\suptext{1}
% \and
% Xuan Wang\suptext{1,}\thanks{Corresponding author}
% \and
% Si Shi\suptext{1} 
% \and
% Xule Wang\suptext{1,}\footnotemark[1]
% \and
% Xinbo Gao\suptext{2} 
% \and
% \suptext{1}Xidian University  $^2$Meituan \\
% $^3$Chongqing University of Posts and Telecommunications\\
% }
\author{
\quad {Chaohua Shi$^{*,1,2}$ ~~~ Xuan Wang$^{2}$ ~~~ Si Shi$^{1,2}$ ~~~ Xule Wang$^{\dag,2}$} \vspace{0.1cm}\\
~~~~~~ Mingrui Zhu$^{1}$~~~ Nannan Wang$^{\dag,1}$~~~ Xinbo Gao$^{1,3}$\vspace{0.1cm}\\
\qquad ~~~~ {$^1$ Xidian University ~~~ $^2$ Meituan Inc.} ~~~~~~~ \vspace{0.1cm}\\
\quad ~~~~ {$^3$ Chongqing University of Posts and Telecommunications} ~~~~~~~ \vspace{0.1cm}\\
\quad \; ~~~~~\small{{chshi2004@gmail.com, sishi@stu.xidian.edu.cn, xbgao@cqupt.edu.cn}} ~~~~~~~ \vspace{0.1cm}\\
~~~~\small{{\{wangxuan39, 
wangxule\}@meituan.com, \{mrzhu, nnwang\}@xidian.edu.cn}}}
\maketitle

%%%%%%%%% ABSTRACT
\begin{abstract}
\blfootnote{$^*$ Work done during the students’ internships at Meituan Inc.}
\blfootnote{$^\dag$ Corresponding author}
Food image composition requires the use of existing dish images and background images to synthesize a natural new image, while diffusion models have made significant advancements in image generation, enabling the construction of end-to-end architectures that yield promising results. However, existing diffusion models face challenges in processing and fusing information from multiple images and lack access to high-quality publicly available datasets, which prevents the application of diffusion models in food image composition. In this paper, we introduce a large-scale, high-quality food image composite dataset, \textbf{FC22k}, which comprises 22,000 foreground, background, and ground truth ternary image pairs. Additionally, we propose a novel food image composition method, \textbf{Foodfusion}, which leverages the capabilities of the pre-trained diffusion models and incorporates a Fusion Module for processing and integrating foreground and background information. This fused information aligns the foreground features with the background structure by merging the global structural information at the cross-attention layer of the denoising UNet. To further enhance the content and structure of the background, we also integrate a Content-Structure Control Module. Extensive experiments demonstrate the effectiveness and scalability of our proposed method.
\end{abstract}

%%%%%%%%% BODY TEXT
\section{Introduction}

Food image composition aims to seamlessly integrate input foreground food images with background images to create high-quality, well-composed synthesized images. This task has numerous applications, including digital advertising, food photography, and augmented reality, which can significantly enhance consumer shopping experiences and reduce the costs associated with producing promotional posters, recipe images and advertisements for catering businesses~\cite{min2023large}.

Food image composition faces two significant challenges. Firstly, large-scale, high-quality, publicly available datasets must be tailored for image generation. Existing datasets, like ETH Food-101~\cite{bossard2014food}, Vireo Food-172~\cite{chen2016deep}, and ISIA Food-500~\cite{min2020isia}, are primarily designed for recognition tasks and are insufficient for developing advanced generative models. Although the Food2k~\cite{min2023large} dataset supports tasks such as recognition and cross-modal recipe retrieval, it is inadequate for composition due to its low image quality and unclear foreground-background relationships.
Secondly, achieving realistic and natural synthesized images remains difficult.
Some generative models can cover the generation of some foods, but due to the poor performance in representing the physical laws of the real world, it is difficult to generate images for scenes with multiple dishes or specified backgrounds. Previous image composition methods~\cite{lin2018st,chen2019toward,lu2023tf} often split the task into subtasks like object placement~\cite{dvornik2018modeling,kikuchi2019regularized,zhang2020learning}, image blending~\cite{wu2019gp,zhang2020deep}, and harmonization~\cite{tsai2017deep,cun2020improving,cong2020dovenet}, which rely heavily on each subtask's performance. This approach often results in inconsistencies that degrade image quality. Additionally, these methods are unsuited for food images, as they fail to preserve detailed features such as texture, colors, patterns, and lines~\cite{niu2021making}.

To address these challenges, we launch FC22k, a large-scale, high-quality food image composition dataset comprising 22,000 foreground, background, and GT triplet image pairs. This dataset provides a solid foundation for training and evaluating food image composition models, ensuring diverse and comprehensive coverage of various synthetic scenes. Compared with existing datasets, FC22k is specifically designed for food image composition tasks, filling the dataset gap for this task. At the same time, we also conducted rigorous data cleaning, iterative annotation, and multiple professional checks to ensure the quality of the data.

Based on this dataset, we propose a novel method, Foodfusion, designed explicitly for food image composition. Our approach leverages a large-scale pre-trained latent diffusion model and incorporates two key modules: the Fusion Module (FM) and the Content-Structure Control Module (CSCM). The Fusion Module utilizes a fusion encoder to encode the foreground and background images into a unified embedding space with multi-scale and spatial awareness. The designed fusion mapping network then merges these embeddings into a unique fused embedding. During this fusion process, the cross-attention layer in the diffusion model UNet ensures a harmonious integration of foreground and background elements. Additionally, the Content-Structure Control Module maintains pixel-level content consistency with the background throughout the fusion process. Extensive experiments conducted on the FC22k dataset demonstrate the effectiveness and scalability of the proposed method. In summary, our contributions are:

\begin{itemize}
    \item FC22k,a comprehensive and high-quality dataset designed for food image composition, is introduced. Which can also be utilized for food image generation tasks.
    %\{We introduce FC22K}, a comprehensive and high-quality dataset designed for food image composition, which can also be utilized for food image generation tasks.
    \item A novel method id designed
    %\{We propose a novel method} 
    for food image composition, Foodfusion, which is the first approach utilizing a latent diffusion model specifically designed to address the challenges of food image composition.
    \item The effectiveness and scalability of our method is demonstrated
    %\We demonstrate the effectiveness and scalability of our method
    through extensive experiments, establishing a new benchmark for food image composition tasks.
\end{itemize}

\section{Related Work}
\subsection{Diffusion-based Image Generation}

Recently, diffusion models~\cite{ho2020denoising,song2020denoising,nichol2021improved,rombach2022high} have been extensively employed in various image generation tasks, including text-to-image generation~\cite{podell2023sdxl,ramesh2022hierarchical,saharia2022photorealistic}, image editing~\cite{cao2023masactrl,zhang2023sine,chefer2023attend,bodur2024iedit}, controllable generation~\cite{zhang2023adding,ma2024directed,mou2024t2i,zhao2024uni}, and subject-driven generation~\cite{jia2023taming,ruiz2023dreambooth,zhang2024ssr,ruiz2024hyperdreambooth}. 
%\Approaches based on the diffusion paradigm and their subsequent adaptations have become the preferred choice for visual information processing tasks.

With its powerful generative capability, some approaches employ diffusion models to perform multiple subtasks simultaneously (such as object placement, image blending, image harmonization, and view synthesis) to develop a unified model capable of generating synthetic images directly. These methods regenerate foreground objects rather than restrictively adjusting them and can be categorized into two types: text-guided~\cite{feng2022training,liu2022compositional,chefer2023attend} and image-guided~\cite{li2023gligen,yang2023paint,song2023objectstitch,lu2023tf}. Text-guided composition involves specifying foreground objects solely based on text prompts, allowing for composition without restricting the appearance of objects as long as their semantics match the prompts. Despite significant successes with text-conditional diffusion models, they often encounter semantic errors~\cite{feng2022training,ramesh2022hierarchical}, mainly when text prompts involve multiple objects. These errors include attribute leakage, attribute swapping, object omission, and generating images deviating significantly from user intentions.

%With its powerful generative capability, some approaches employ diffusion models to perform multiple subtasks simultaneously (such as image blending, image harmonization, and view synthesis) to develop a unified model capable of generating synthetic images directly. These methods regenerate foreground objects rather than restrictively adjusting them, thus termed generative image synthesis methods. They can be categorized into two types: text-guided~\cite{feng2022training,liu2022compositional,chefer2023attend} and image-guided~\cite{li2023gligen,yang2023paint,song2023objectstitch,lu2023tf}. Text-guided synthesis involves specifying foreground objects solely based on text prompts, allowing for composition without restricting the appearance of objects as long as their semantics match the prompts. Despite significant successes with text-conditional diffusion models, they often encounter semantic errors~\cite{feng2022training,ramesh2022hierarchical}, mainly when text prompts involve multiple objects. These errors include attribute leakage, attribute swapping, and object omission, generating images that deviate significantly from user intentions, necessitating prompt engineering~\cite{witteveen2022investigating} to achieve desired outcomes.

In contrast, image-guided composition integrates specific foreground objects and backgrounds from user-provided photos with text prompts~\cite{li2023gligen,yang2023paint,lu2023tf}. However, these methods face challenges in processing and merging information from multiple images, mainly when substantial differences exist between foreground objects and backgrounds~\cite{lu2023tf}. Additionally, these methods are not well-suited for food image composition due to two primary reasons: they fail to preserve detailed features of the foreground and the datasets utilized in previous image composition tasks are not food-related, highlighting the lack of high-quality, large-scale food image composition datasets.

\subsection{Image Composition}

Image composition~\cite{zhan2020adversarial,niu2021making,song2022objectstitch} has been a prominent research area in computer vision, focusing on combining one image's foreground with another's background to create a cohesive composite image. Image composition involves integrating multiple visual elements from different sources to construct a new image, which is a typical operation in image editing. Traditional methods typically divide this task into several subtasks~\cite{niu2021making}, such as object placement~\cite{dvornik2018modeling,kikuchi2019regularized,zhang2020learning}, image blending~\cite{wu2019gp,zhang2020deep}, and harmonization~\cite{tsai2017deep,cun2020improving,cong2020dovenet,xue2022dccf}. For instance, object placement methods model object relationships to position them appropriately within the scene. Image blending techniques aim to seamlessly integrate foreground objects with the background, ensuring consistent texture and lighting. Harmonization methods adjust the appearance of the foreground to match the background in terms of color, brightness, and texture. Although these methods offer practical solutions to the image composition task, they often need help to preserve the fine details required for the foreground image. Meanwhile, they are heavily dependent on the performance of the individual subtask models.

\section{Dataset Construction}

\begin{figure*}[t]
\centering
\includegraphics[width=0.91\textwidth]{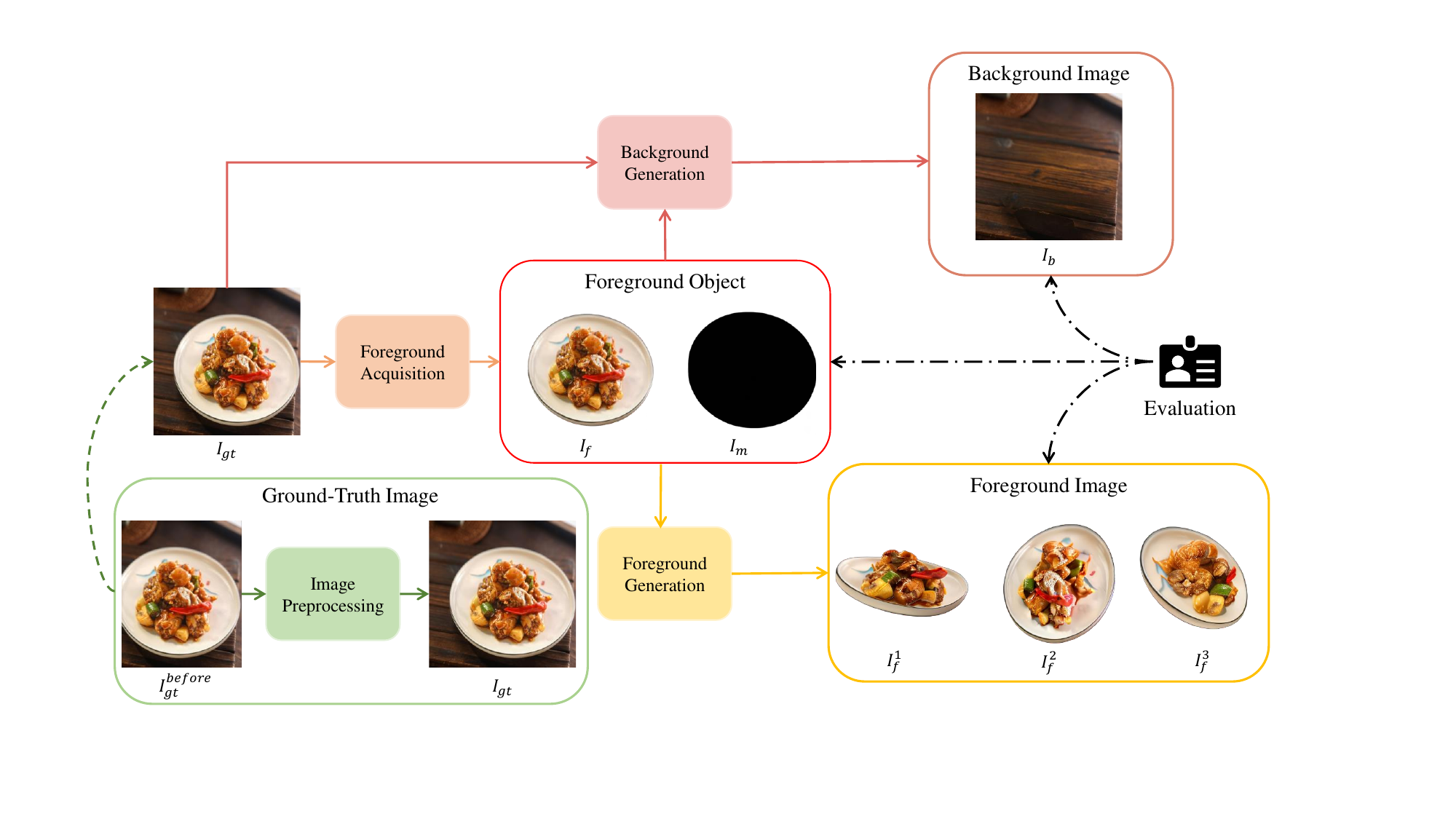} 
\caption{The illustration of our dataset (FC22k) construction process. Starting with a ground truth ($I_{gt}$) image containing a clear foreground ($I_{f}$) and background ($I_{b}$), our automated process generates multiple data pairs with different foregrounds but the same background, along with their corresponding GT images.}
\label{fig:dataset}
\end{figure*}

\begin{figure}[t]
\centering
\includegraphics[width=0.9\columnwidth]{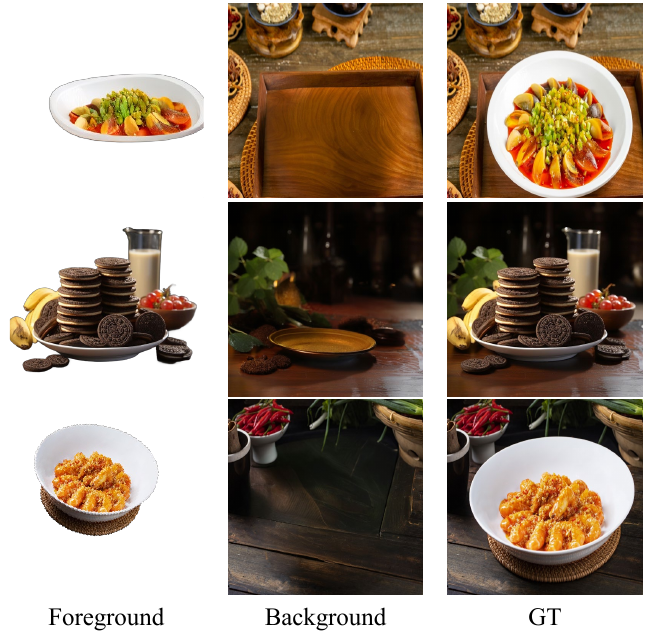} 
\caption{Some samples from FC22k dataset.}
\label{fig:dataset_show}
\end{figure}

This section details the automated construction process of the food composition dataset FC22k, as illustrated in Fig.~\ref{fig:dataset}. The process comprises five main stages: image preprocessing, foreground acquisition, foreground generation, background generation, and evaluation. Through these stages, we have created a large-scale, high-quality dataset for food image composition, consisting of 22,000 foreground ($I_{f}$), background($I_{b}$), and ground truth (GT) triplet image pairs.

\subsection{Image Preprocessing}

Image preprocessing aims to enhance the quality of ground truth (GT) images $I_{gt}$ to facilitate subsequent processing stages. Since GT images are primarily sourced online, their quality varies significantly. Therefore, it is essential to first screen and enhance these images using various image processing techniques~\cite{chen2023learning}, such as resolution screening, image denoising, deblurring, and watermark removal. For GT images where the foreground objects occupy a substantial proportion, we utilize SDXL~\cite{podell2023sdxl} to expand the images appropriately. This ensures that the foreground area occupies a reasonable portion of the image, aiming to cover a wide range of food categories.

\subsection{Composited Image Pairs Generation}

Obtaining the corresponding foreground and background from the GT image involves three main stages: foreground acquisition, foreground generation, and background generation. In the foreground acquisition stage, we extract the foreground $I_f$ and corresponding mask $I_m$ from $I_{gt}$ through multiple segmentations using SAM~\cite{kirillov2023segment} and RMBG1.4~\cite{qin2022highly}. The extracted foreground and mask are then processed in the foreground generation stage, where images from different perspectives are generated using a 3D generation model~\cite{shi2023zero123++}. This approach better simulates real-world scenarios instead of merely increasing foreground diversity through affine transformations. For the background generation stage, we input $I_{gt}$ and the $I_m$ obtained from the foreground segmentation into SDXL~\cite{podell2023sdxl}, using it to perform an inpainting task that repaints the foreground area, thus generating the corresponding background image $I_b$.

\subsection{Evaluation}

Through the foreground and background generation stages, we produce numerous synthetic images. However, due to the inherent randomness of the generation models, some generated images may lack realism. To address this, we employ an image quality score model~\cite{kong2016photo} to filter out unrealistic images. To further refine the dataset, we use the real images identified by the quality score model as positive samples and the remaining images as negative samples to train a binary classification network. This model provides an additional layer of evaluation for the generated images.

Despite these automated measures, some unrealistic images may remain. Therefore, we conduct a manual review to remove any remaining unrealistic images. We successfully constructed the FC22k food image composition dataset through this comprehensive process. Fig.~\ref{fig:dataset_show} illustrates some examples from this dataset.

\begin{figure*}[t]
\centering
\includegraphics[width=0.95\textwidth]{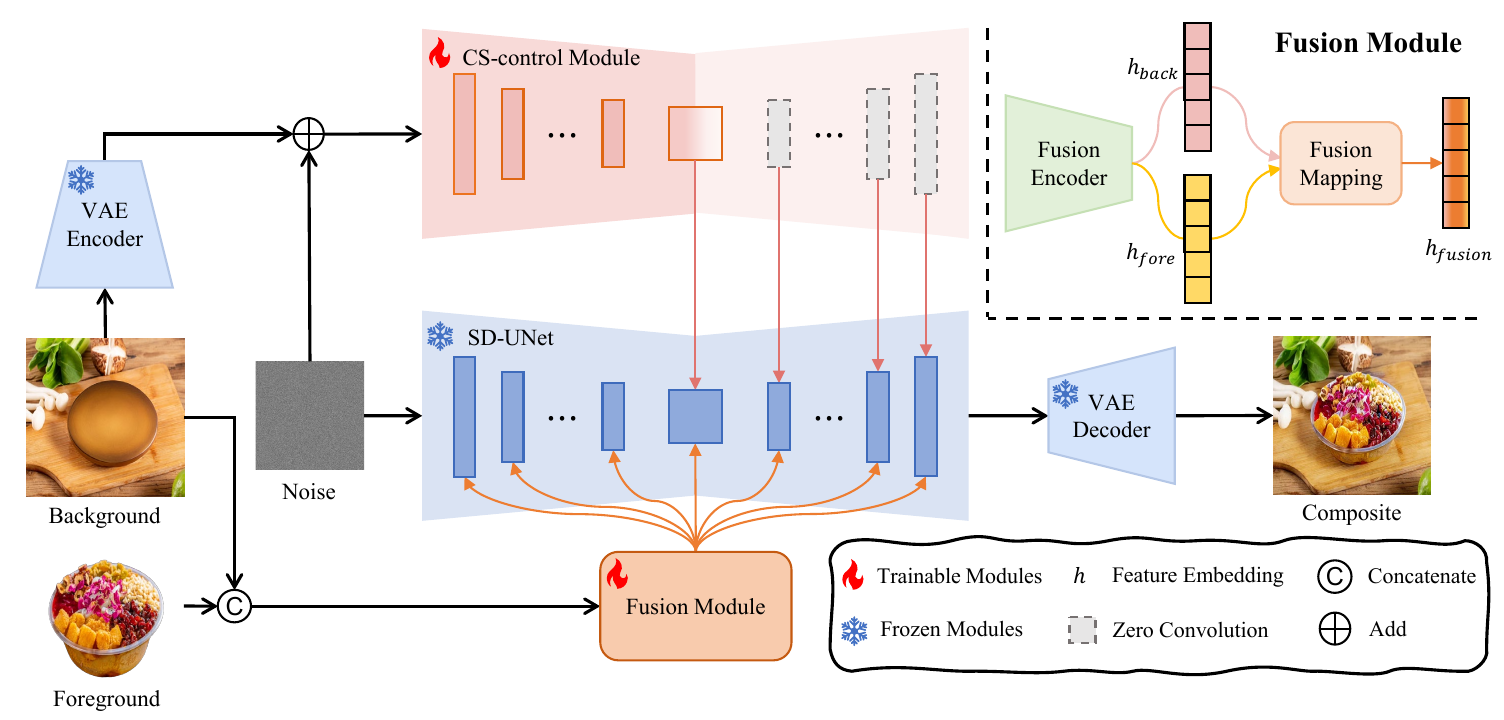} % Reduce the figure size so that it is slightly narrower than the column.
\caption{Ovearview of our proposed Foodfusion model. Given a foreground food image $I_f$ and a background $I_b$, Foodfusion effectively processes and merges them. By automatically adjusting the foreground's size, angle, and position, it seamlessly integrates $I_f$ with $I_b$ to create a high-quality composite image $I_c$.}
\label{fig:framework}
\end{figure*}

\section{Method}

In this section, we introduce Foodfusion, depicted in Fig.~\ref{fig:framework}, which seamlessly integrates an input foreground food image $I_f$ into a user-provided background $I_b$ by automatically adjusting the foreground's size, angle, and position to create a high-quality, well-placed, and well-composed synthetic image $I_c$, utilizing a large-scale pre-trained latent diffusion model and two key modules—the Fusion Module, which harmonizes foreground and background within the stable diffusion model, and the Content-Structure Control Module, which ensures pixel-level content consistency with the background throughout the fusion process.

% \begin{figure}[t]
% \centering
% \includegraphics[width=0.9\columnwidth]{image/Fusion_Module.pdf} 
% \caption{Fusion Module.}
% \label{fig:fusion_module}
% \end{figure}

\subsection{Fusion Module}

The Fusion Module integrates the foreground $I_f$ and background $I_b$ into a unified embedding space with multi-scale and spatial perception, subsequently fusing this information and feeding the resulting fused embedding $h_{fusion}$ into the cross-attention layer of the Stable Diffusion~\cite{rombach2022high} denoising UNet. This module comprises three main components: the Fusion Encoder $E_f$, the Fusion Mapping Network $M_f$, and the cross-attention layer.

\textbf{Fusion Encoder}: The Fusion Encoder $E_f$ takes $I_f$ and $I_b$ as input. However, food foregrounds often contain complex details such as texture, colors, patterns, and lines. These details are often subtle and precise, making extracting and encoding foreground images challenging. We use the pre-trained CLIP~\cite{radford2021learning} image encoder to extract features from different layers and concatenate them along the feature dimension to address this issue. This encoding method can capture fine-grained details and spatial information at different resolutions, thereby encoding the foreground and background into a unified embedding space with multi-scale and spatial awareness.

\begin{equation}
\begin{gathered}
h_{fore}=E_f\left(I_b\right)\text {, } \\
h_{back}=E_f\left(I_f\right)\text {, }
\end{gathered}
\end{equation}
where $h_{fore}$ and $h_{back}$ denote the feature embeddings of foreground and background respectively, which are in the same embedding space.

\begin{figure}[t]
\centering
\includegraphics[width=0.98\columnwidth]{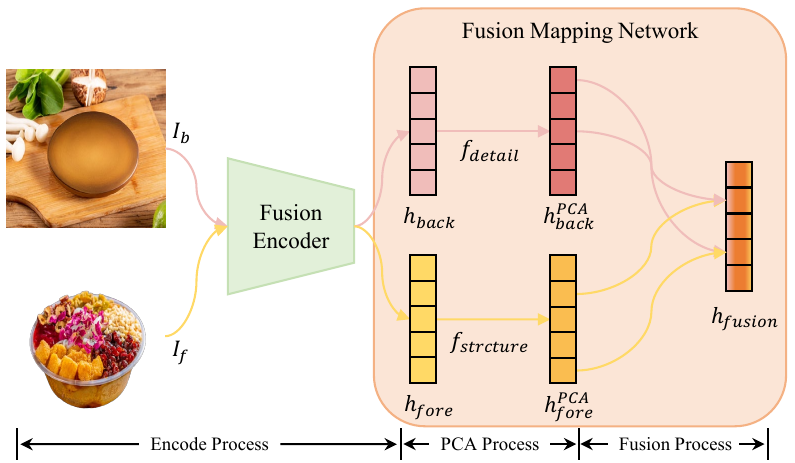} 
\caption{The illustration of the Fusion Module. It can effectively process and fuse foreground and background images.}
\label{fig:fusion_module}
\end{figure}

\textbf{Fusion Mapping Network}: To facilitate the interaction between foreground and background information, we designed a fusion mapping network $M_f$ to map the foreground embedding $h_{fore}$ and background embedding $h_{back}$ in the same latent space into a unique fused embedding $h_{fusion}$. As shown in Fig.~\ref{fig:fusion_module}{}, this network extracts essential details from the foreground embedding and the most relevant structural position features from the background embedding in a PCA-like manner. It then fuses and maps these features back to the original embedding space. This process preserves the spatial relationship between foreground and background elements, ensuring that the foreground can be adaptively adjusted based on the structural information of the background.

\begin{equation}
h_{fusion}=M_f\left(h_{fore}, h_{back}\right)\text {, }
\end{equation}
where $h_{fusion}$ denotes the unique fused embedding, which is in the same embedding space as $h_{fore}$ and $h_{back}$.

\textbf{Cross-attention layer}: The fused embedding is fed into each cross-attention layer in the Stable Diffusion~\cite{rombach2022high} denoising UNet. And, the cross-attention layer implements $\operatorname{Attention}(Q, K, V)=\operatorname{softmax}\left(\frac{Q K^T}{\sqrt{d}}\right) \cdot V$.  This cross-attention mechanism matches the foreground to the appropriate position in the background. It ensures a seamless integration of foreground and background elements by dynamically adjusting the weighting of each component based on its relevance to the overall composition. The specific process is as follows:

\begin{equation}
Q=W_Q^{i} \cdot \varphi_i\left(z_t\right), K=W_K^{i} \cdot h_{fusion}, V=W_V^{i} \cdot h_{fusion},
\end{equation}
where $\varphi_i$ denotes  the $i^{th}$ (flattened) intermediate feature of the denoising UNet. $W_Q^{i}$, $W_K^{i}$ and $W_V^{i}$ are learnable projection matrices.

\subsection{Content-Structure Control Module}

In order to maintain the structural consistency of the background during the feature transmission process of the Stable Diffusion~\cite{rombach2022high} denoising UNet, we use a Content-Structure Control Module (CSCM). Its goal is to maintain the consistency of the background and the synthesized image in the non-foreground area. Fig.~\ref{fig:framework} shows the architecture of our CSCM. It is essentially the same as the Stable Diffusion~\cite{rombach2022high} denoising UNet. By taking the background encoded by the VAE encoder $\mathcal{E}$ as input,  the resulting content-structure features are integrated into the stable diffusion denoising UNet like ControlNet~\cite{zhang2023adding}. Finally, to prevent the text from interfering with the background content-structure features, we use the text embedding of the empty text as the input of the cross-attention layer of this module.

\begin{equation}
F_{back}^{i} = CSCM\left(\mathcal{E}(I_b)\right)\text {, }
\end{equation}
where $F_{back}^{i}$ denotes the $i^{th}$ intermediate representation of the denoising UNet implementing $\epsilon_\theta$.

\subsection{Training Procedure}

We employ the original diffusion loss~\cite{rombach2022high} to train our Foodfusion model on the FC22k dataset. This loss function ensures that the synthesized image retains the essential features of the original foreground and background images while achieving seamless blending. By integrating the Fusion Module and the Content-Structure Control Module, the loss function can be formulated as follows:

\begin{equation}
L=\mathbb{E}_{(I_{gt}, I_f, I_b), \epsilon, t}\left[\left\|\epsilon-\epsilon_\theta\left(z_t, t, h_{fusion}, F_{back}\right)\right\|_2^2\right] \text {, }
\end{equation}
where $z_t$ is a noisy image latent constructed by adding noise $\epsilon \in \mathcal{N}(\mathbf{0}, \mathbf{1})$ to the image latents $z_0=\mathcal{E}(I_{gt})$.

In addition, we incorporate various enhancement techniques into our training process to adapt to real-world scenarios and achieve successful food image composition. These enhancements are categorized into foreground enhancement and background enhancement. Foreground enhancement includes structural modifications, such as introducing appropriate distortions (e.g., noise, blur, or pixel loss) and performing affine transformations. These techniques aim to increase the complexity and diversity of the foreground samples. Background enhancement enables our model to adapt to different background sizes encountered in real-world scenes.

\begin{figure*}[t]
\centering
\includegraphics[width=\textwidth]{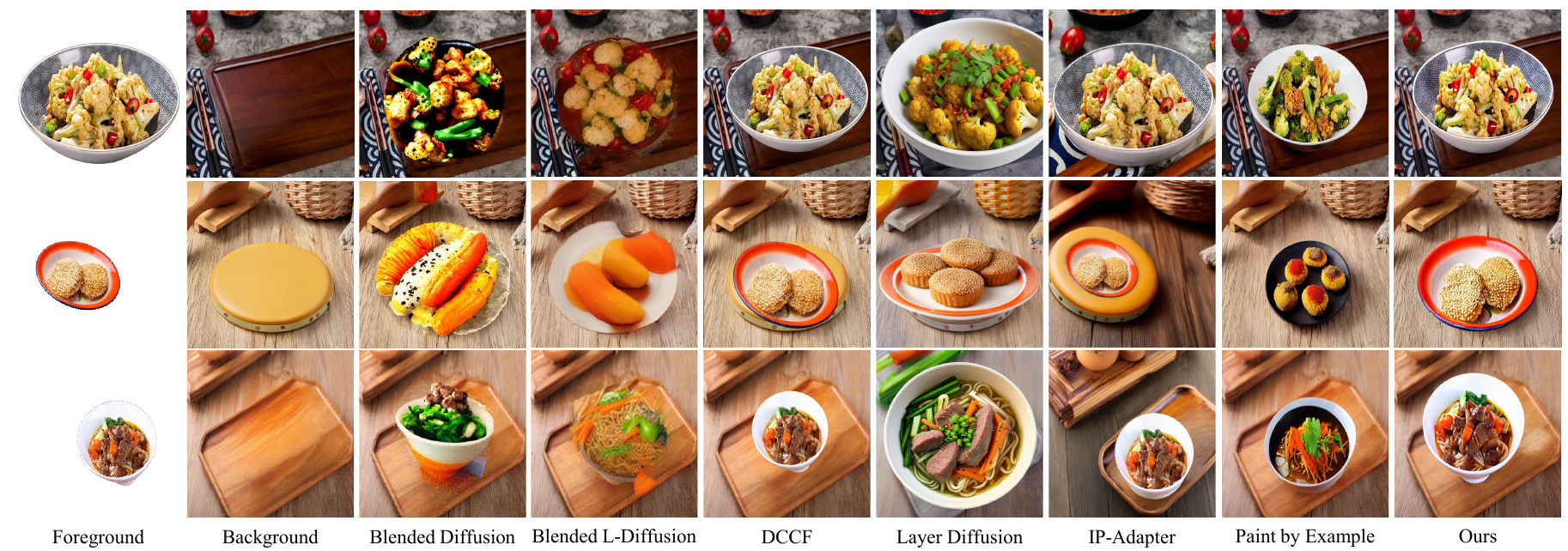}
\caption{Qualitative comparison with other methods. Our method effectively fuses foreground $I_f$ and background $I_b$ information without requiring additional positional data, such as masks, to generate high-quality food composite image $I_c$.}
\label{fig:compare}
\end{figure*}

\section{Experiments}

In this section, we evaluate the effectiveness of the proposed Foodfusion method using the newly introduced FC22k dataset. We detail the experimental setup, including dataset specifications, evaluation metrics, and implementation procedures. We comprehensively evaluate our method and discuss its potential for practical applications.

\subsection{Experimental Setup}

\subsubsection{Dataset} The FC22k dataset consists of 22,000 triplets, each containing a foreground food image, a background image, and an actual composite image. This dataset is designed for the task of food image composition and contains a variety of food, background, and synthetic scenes. To ensure the effectiveness of the experiment, we divide the dataset into training $(80\%)$, validation $(10\%)$ and test $(10\%)$ sets.

\subsubsection{Implementation Details} We employ Stable Diffusion V1.5 as the pre-trained diffusion model, updating the network parameters for our proposed fusion module and content structure control module while keeping the rest parameters frozen. Training is conducted on 4 NVIDIA A100 GPUs and the batch size is 12, the initial learning rate 5e-5, and the Adam optimizer with $\beta_1=0.5$ and $\beta_2=0.99$. The training process spans 300 epochs, with early stopping applied based on validation loss. During inference, we use DDIM as the sampler with step size of 30 and guidance scale of 1.5.
%Our Foodfusion method is implemented using PyTorch. We employ Stable Diffusion V1.5 as the pre-trained diffusion model, updating the network parameters for our proposed fusion module and content structure control module on the FC22K dataset while keeping the rest of the model parameters frozen. Training is conducted on 4 NVIDIA A100 GPUs with a batch size of 12, an initial learning rate 5e-5, and the Adam optimizer with $\beta_1=0.5$ and $\beta_2=0.99$. The training process spans 300 epochs, with early stopping applied based on validation loss. During inference, we use DDIM as the sampler with a step size of 30 and a guidance scale of 1.5.

\subsubsection{Evaluation Metrics} Our goal is to blend the foreground into the background naturally while preserving the key features of the foreground. To evaluate the quality of the generated images, we use three indicators on the FC22k test set: (1) PSNR, which measures image quality by comparing each composite image with the ground truth (GT) and averaging the results. (2) LPIPS assesses the visual similarity between the composite and GT images using the same method as PSNR. (3) User Study, where 50 participants selected the best-quality image from 20 sets, with images presented randomly to gather subjective evaluations.

\subsection{Comparisons}

Considering the absence of prior research on food image composition, we selected six relevant methods for comparative analysis: (1) Blended Diffusion~\cite{Avrahami_2022_CVPR} utilizes CLIP-derived gradient information to guide its diffusion model sampling, supplemented by GPT4 for foreground representation via textual hints. (2) Blended Latent Diffusion~\cite{avrahami2023blended}, akin to (1), employs pre-trained Stable Diffusion (SD). (3) DCCF~\cite{xue2022dccf} is recognized as a cutting-edge image harmonization technique. (4) Layer Diffusion~\cite{zhang2024transparent} is an enhancement of SD that enables image generation with transparency and multiple transparent layers. (5) IP-Adapter~\cite{ye2023ip}, an established extension of SD capable of injecting tailored conditional guidance into the generation process. (6) Paint by Example~\cite{yang2023paint}, an advanced image editing method leveraging SD to intelligently replace masked areas in original images based on exemplar images.

\subsubsection{Results $\&$ Analysis}

Fig.~\ref{fig:compare} visually compares our proposed method, Foodfusion, with other related methods. These methods are categorized into text-guided and image-guided image compositions. Blended Diffusion and Blended Latent Diffusion are text-guided diffusion model-based blended methods. We convert the foreground into corresponding text prompts using GPT-4 in these methods (the detail text is shown in \textbf{Supplementary Material}). While they can generate foreground objects relevant to the text prompts, these methods often lack realism and are incompatible with the background, resulting in noticeable artefacts at the edges. Layer Diffusion, another text-based diffusion model, produces more realistic results but struggles to preserve the user-specified foreground characteristics due to the inherent differences between text and image representations. This limitation is problematic for food image composition, which demands preserving fine details in the foreground food and its proper integration into the background image for higher commercial value.

% Another class of methods is image compositions based on image guidance. DCCF is the most advanced image coordination method, but its results are almost identical to the foreground and inconsistent with the background. The underlying reason is that, in most cases, the appearance of the foreground cannot be directly matched with the background. A good image synthesis model should automatically transform the foreground shape, size, or posture to adapt to the background. IP-Adapter fixes the foreground area and redraws the background information in the non-foreground area, so the background of the result it generates will change significantly. Paint by Example can reasonably merge and replace the mask area of the background according to the foreground. However, it cannot effectively preserve the detailed features of the foreground, such as texture, colors, patterns, and lines.

Another class of methods is image compositions based on image guidance. DCCF is the most advanced image coordination method, but its results are almost identical to the foreground and inconsistent with the background. The underlying reason is that, in most cases, the appearance of the foreground cannot be directly matched with the background. A good image composition model should automatically transform the foreground shape, size, or posture to adapt to the background. IP-Adapter fixes the foreground area and redraws the background information in the non-foreground area, so the background of the result it generates will change significantly. Paint by Example can reasonably merge and replace the mask area of the background according to the foreground. However, it cannot effectively preserve the detailed features of the foreground, and the foreground in the composite image $I_c$ needs to be more consistent with the given foreground $I_f$.

In contrast, our method not only effectively maintains the fine details of the foreground but also automatically adjusts the foreground's shape, size, or posture according to the background. Additionally, our approach matches the foreground to the appropriate background position without needing extra positional data like masks, which other methods require. The quantitative comparison results in Table~\ref{tab:metric} also further illustrate the superiority of our method, which achieves the best performance in objective and subjective evaluation metrics. More visual results could be seen in \textbf{Supplementary Material}

\begin{table}[]
\caption{Quantitative comparison of different methods. Our method achieves state-of-the-art performance on both objective and subjective metrics. The best results are in \textbf{bold} and the second best results are marked with an \underline{underline}.}
\label{tab:metric}
\centering
\begin{tabular}{lccc}
\toprule[1.5pt]
Method              & PSNR~$\uparrow$ & LPIPS~$\downarrow$           & User Study~$\uparrow$ \\ \midrule
Blended Diffusion   & 8.96 & 0.4642          & 2$\%$        \\
Blended L-Diffusion & 10.54 & 0.4225          & 3$\%$        \\
DCCF                & 11.47 & 0.3801          & 10$\%$        \\
Layer Diffusion     & 12.38 & 0.3927          & 3$\%$        \\
IP-Adapter          & \underline{16.58} & \underline{0.3619}          & \underline{15$\%$}         \\
Paint by Example    & 14.21 & 0.3711          & 12$\%$         \\
Ours                & \textbf{22.05} & \textbf{0.2501} & \textbf{55$\%$}        \\ \bottomrule[1.5pt]
\end{tabular}
\end{table}

\begin{figure}[t]
\centering
\includegraphics[width=\columnwidth]{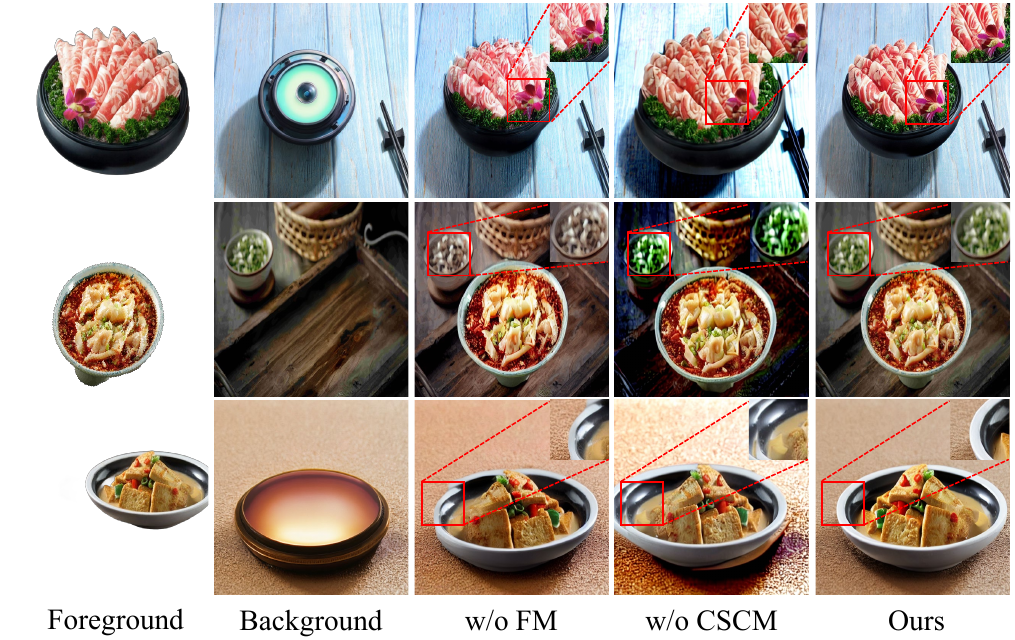} 
\caption{Qualitative ablation results. We verify the importance of the Fusion Module (FM) and the Content-Structure Control Module (CSCM) in our method in different experimental settings.}
\label{fig:ablation}
\end{figure}

\subsection{Ablation Study}

To achieve high-quality food image composition, our method leverages pre-trained Stable Diffusion and employs two key modules: the Fusion Module (FM) and the Content-Structure Control Module (CSCM). These modules effectively process and fuse foreground and background information. In this subsection, we validate their significance through various experimental setups: (1) We replace the fusion module in our method with the original image encoder in CLIP for training, directly modifying the text-guided Stable Diffusion-based in-painting model by using images instead of text as conditional signals. (2) To assess the enhancement of the content-structure control module on background content structure information, we conduct training without this module.

% The results of the ablation experiment are illustrated in Fig.~\ref{fig:ablation}. Omitting our designed Fusion Module results in composite images that fail to retain the fine details of the foreground food (first row) and introduce blur and artefacts at the edges of the foreground (third row). Additionally, the fusion module helps preserve the correct background color in the composite images. Regarding the Content-Structure Control Module, the results indicate that this module significantly enhances the content consistency between the composite image and the background in pixel level, eliminating artefacts at the foreground edges (third row). When fully implemented, our method achieves a coherent composition of foreground and background, preserving the critical information from both types of images in the final composite.
The results of the ablation experiment are illustrated in Fig.~\ref{fig:ablation}. Omitting our designed Fusion Module results in composite images that fail to retain the fine details of the foreground food (first row) and introduce blur and artefacts at the edges of the foreground (third row). This is because the original image encoder in CLIP cannot capture the detailed features of the foreground well and cannot fuse image information. Additionally, the fusion module helps preserve the correct background color in the composite images. Regarding the Content-Structure Control Module, the results indicate that this module significantly enhances the content consistency between the composite image and the background at the pixel level, eliminating artefacts at the foreground edges (third row) for injecting additional background features into the  Stable Diffusion denoising Unet. When fully implemented, our method achieves a coherent composition of foreground and background, preserving the critical information from both types of images in the final composite.

\subsection{Expansion Discussion}

\begin{figure}[t]
\centering
\includegraphics[width=0.95\columnwidth]{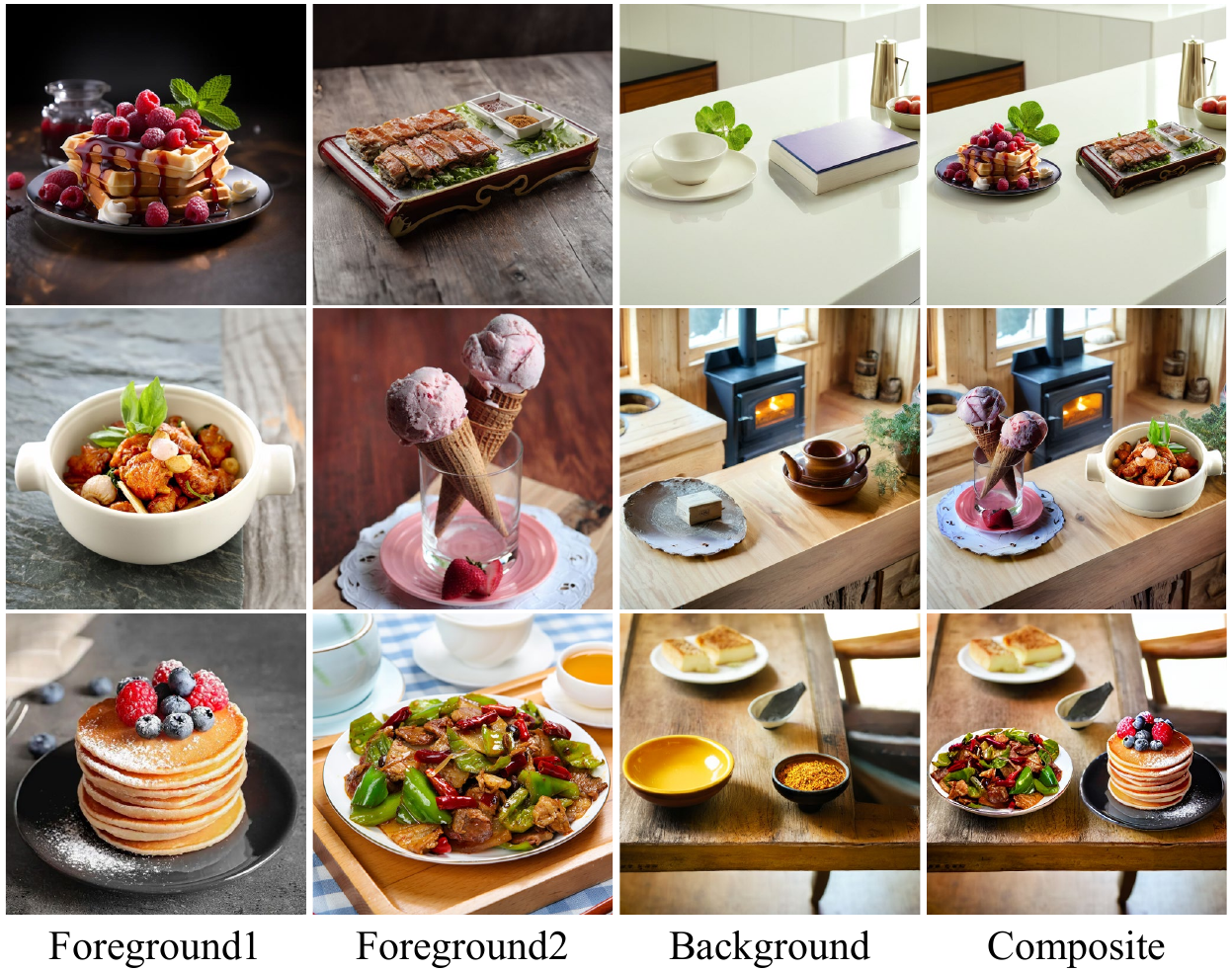} 
\caption{Extended experimental results on complex food image composition.}
\label{fig:complex_food}
\end{figure}

% \begin{figure}[t]
% \centering
% \includegraphics[width=0.8\columnwidth]{image/Poeple.pdf} 
% \caption{Extended experimental results on different image composition task.}
% \label{fig:people}
% \end{figure}

In this subsection, we discuss the scalability of our method and demonstrate its generalization capabilities to more complex food image composition scenarios and different image composition task.

Fig.~\ref{fig:complex_food} illustrates the performance of our method in a complex food image composition scenario involving multiple foreground images with some interference elements (background within the foreground images). Our approach does not require extensive modifications; it only necessitates adding additional image branches within the Fusion Module to achieve composites guided by multiple foreground images. This demonstrates the robust scalability of our method, which can adapt to more complex application scenarios with minimal adjustments. The generalizability of our method to different image composition tasks is discussed in the \textbf{Supplementary Material}.

% Fig.~\ref{fig:people} illustrates the performance of our method across various image composition scenarios, using familiar real-world portrait images as examples. The results demonstrate that our proposed method can effectively composite portrait images, highlighting its strong generalization capabilities and excellent performance in different image composition tasks.

\section{Conclusion}

In this paper, we addressed the challenges of food image composition by introducing a large-scale, high-quality dataset, \textbf{FC22K}, and a novel method called \textbf{Foodfusion}. FC22k, consisting of 22,000 foreground, background, and ground truth image pairs, is specifically designed for food image composition, filling a critical gap in existing datasets. Foodfusion leverages pre-trained diffusion models and incorporates a Fusion Module (FM) and a Content-Structure Control Module (CSCM) to ensure seamless integration of foreground and background elements. Extensive experiments on the FC22k dataset demonstrate the effectiveness and scalability of our method, establishing a new benchmark for food image composition tasks. Our results show significant improvements in image quality and consistency compared to previous methods, which often rely on separate subtasks and need help preserving detailed features like texture and color. Future work will enhance our model's capabilities and expand its applicability to other domains.

%%%%%%%%% REFERENCES
{\small
\bibliographystyle{ieee_fullname}
\bibliography{egbib}
}

\noindent\textbf{\large{Supplementary Material}}

% \begin{document}
\appendix
\section{Preliminaries}
    
\begin{figure*}[t]
\centering
\includegraphics[width=\textwidth]{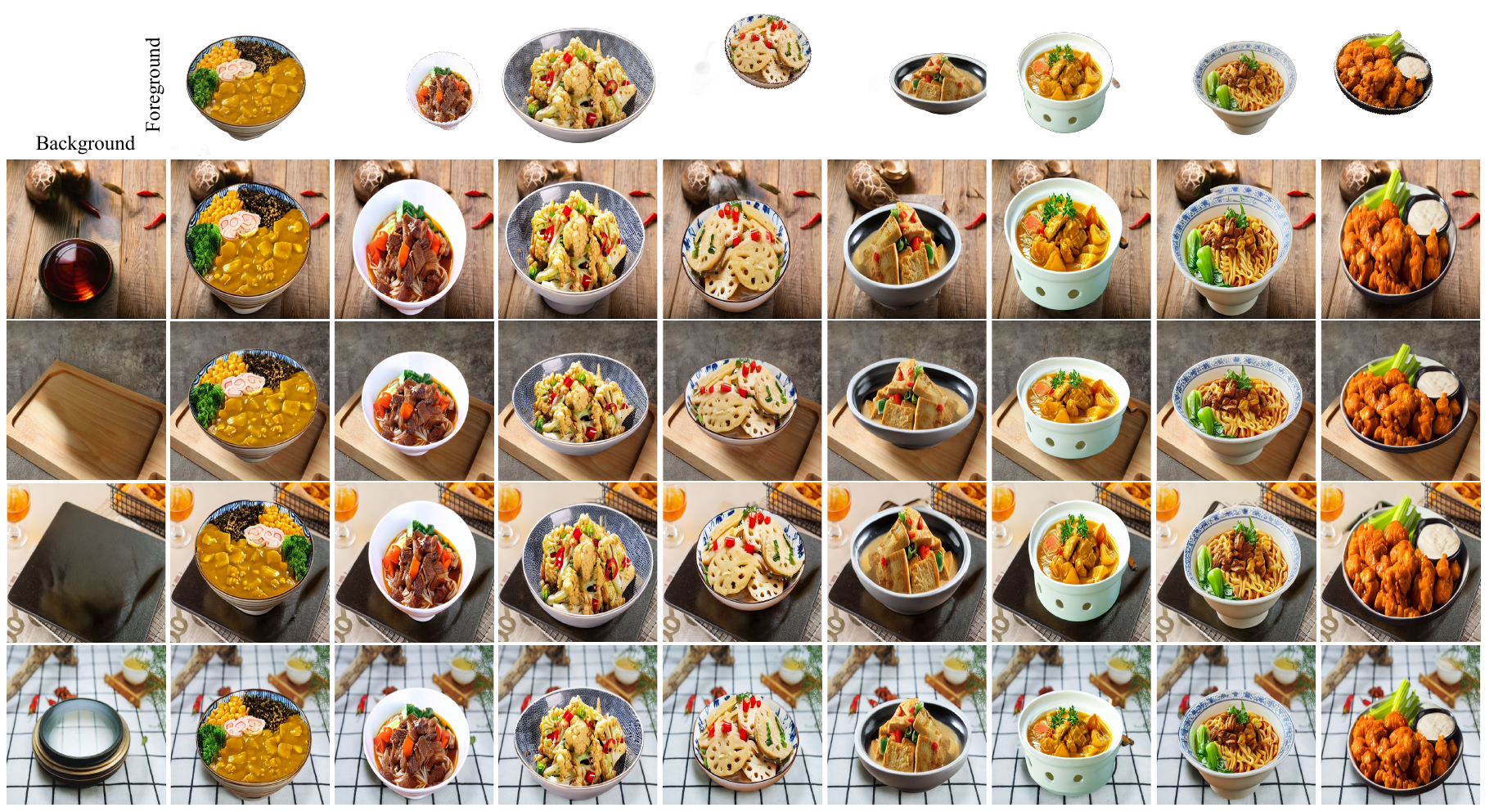} % Reduce the figure size so that it is slightly narrower than the column.
\caption{Visual results of our method (Foodfusion) on different foregrounds and backgrounds. Our method can adaptively adjust the foreground according to the background and generate high-quality synthetic images with reasonable layout without additional guiding information such as text or masks.}
\label{fig:more_result}
\end{figure*}

\begin{figure}[t]
\centering
\includegraphics[width=0.85\columnwidth]{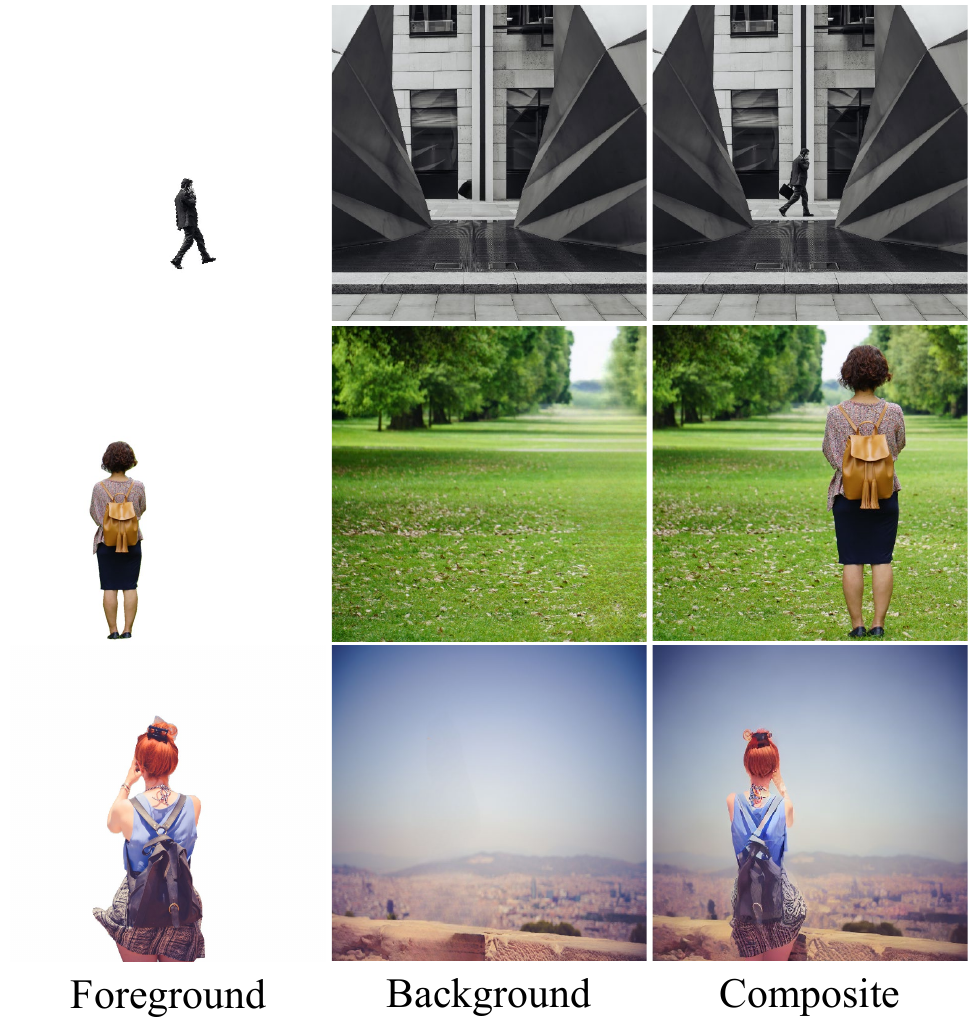} 
\caption{Extended experimental results on different image composition task.}
\label{fig:people}
\end{figure}

In this section, we present the foundational knowledge and key techniques essential for developing our method.
    
\subsection{Diffusion models}
The Diffusion Model (DM)~\cite{ho2020denoising,song2020denoising,nichol2021improved} is a type of generative model that transforms a Gaussian prior $(x_T)$ into a target data distribution $(x_0)$ through an iterative denoising process. The Latent Diffusion Model (LDM)~\cite{rombach2022high} extends this framework by specifically modelling image representations within the latent space of autoencoders. LDM significantly accelerates the sampling process and enhances text-to-image generation by incorporating additional textual conditions. The loss function of LDM is:

\begin{equation}
L_{L D M}(\boldsymbol{\theta}):=\mathbb{E}_{\mathbf{z}_{\mathbf{0}}, t, \boldsymbol{\epsilon}}\left[\left\|\boldsymbol{\epsilon}-\boldsymbol{\epsilon}_{\boldsymbol{\theta}}\left(z_{t}, t, \boldsymbol{\tau}_{\boldsymbol{\theta}}(\mathbf{c}_t)\right)\right\|_2^2\right] \text {, }
\end{equation}
where $z_t$ is the noisy image latent image constructed by adding noise $\epsilon \in \mathcal{N}(\mathbf{0}, \mathbf{1})$ to the image latent image $x_0$, the network $\epsilon_\theta()$ is trained to predict the added noise, and $\tau_\theta()$ refers to the BERT text encoder~\cite{devlin2018bert} used to encode the text description $\mathbf{c}_t$.

Stable Diffusion (SD) is a widely adopted text-to-image diffusion model built upon the Latent Diffusion Model (LDM). Unlike LDM, SD is trained on the extensive LAION dataset~\cite{schuhmann2022laion} and utilizes a pre-trained CLIP text encoder~\cite{radford2021learning} instead of the BERT model~\cite{devlin2018bert}.

\subsection{ControlNet}

ControlNet~\cite{zhang2023adding} is one of the most widely used control modules in current diffusion models. It processes inputs from various modalities as spatial control conditions and directs the diffusion model to generate images according to specific requirements, thereby enabling controllable generation. ControlNet replicates the original U-Net structure as trainable parameters while keeping the parameters of the original U-Net fixed. The entire architecture of ControlNet can be described as follows:

\begin{equation}
\boldsymbol{y}_{\mathrm{c}}=\mathcal{F}(\boldsymbol{z} ; \Theta)+\mathcal{Z}\left(\mathcal{F}\left(\boldsymbol{z}+\mathcal{Z}\left(\boldsymbol{c} ; \Theta_{\mathrm{z} 1}\right) ; \Theta_{\mathrm{c}}\right) ; \Theta_{\mathrm{z} 2}\right) \text {, }
\end{equation}
where $\mathcal{F}$ is denoising UNet, $z$ is image latent, $\Theta$ is the frozen weight of the U-Net and $\Theta_{\mathrm{c}}$ is the trainable copy weight of the U-Net. $\Theta_{\mathrm{z} 1}$ and $\Theta_{\mathrm{z} 2}$ represent two different zero conv layers' parameters respectively.

\subsection{CLIP}

CLIP~\cite{radford2021learning} comprises two core components: an image encoder, denoted as $E_{I}(x)$, and a text encoder, denoted as $E_{T}(t)$. . The image encoder $E_{I}(x)$ transforms an image $x$ of size $\mathbb{R}^{3 \times H \times W}$ (where $H$ is the height and $W$ is the width) into a $d$-dimensional image feature $f_I$ of size $\mathbb{R}^{N \times d}$, with $N$ representing the number of segmented patches. Conversely, the text encoder $E_{T}(t)$ generates a $d$-dimensional text embedding $f_t$ of size $\mathbb{R}^{M \times d}$ from a natural language text $t$, where $M$ corresponds to the number of text tokens. Trained using a contrastive loss function, CLIP can be applied directly to zero-shot image recognition tasks without requiring fine-tuning of the entire model.

\section{Discussion of Data Construction}

The advent of popular applications such as ChatGPT and Stable Diffusion has significantly transformed the AI landscape through the widespread adoption of generative models. Many tasks now rely on these state-of-the-art models to create specialized datasets for training purposes. However, this approach often introduces significant data biases, which can limit the model's generalization capabilities across different tasks~\cite{yang2022pastiche,song2022diffusion,zhang2024stable,lei2024diffusiongan3d}. For instance, employing data generated by Stable Diffusion for a style transfer task may constrain the model to a single style representation inherent in the generated data. In contrast, our FC22k food image composition dataset is built exclusively from authentic, high-quality food images rather than relying on data produced by generative models. Our dataset construction begins with real-world images and leverages advanced generative models to extract the necessary label information for the food image composition task, such as foreground and background. This approach ensures that the target data distribution of the designed model is rooted in actual data, allowing for better simulation of real-world scenarios and minimizing the introduction of biases associated with generative models.

% \begin{figure*}[t]
% \centering
% \includegraphics[width=\textwidth]{image/more_result.pdf} % Reduce the figure size so that it is slightly narrower than the column.
% \caption{Visual results of our method (Foodfusion) on different foregrounds and backgrounds. Our method can adaptively adjust the foreground according to the background and generate high-quality synthetic images with reasonable layout without additional guiding information such as text or masks.}
% \label{fig:more_result}
% \end{figure*}

\section{Experiments}

In this section, we provide additional experimental details to complement the findings presented in the main text. Specifically, we will 1) present the image descriptions generated by GPT-4 that were used in the comparative experiments, 2) showcase additional visual results produced by our method, and 3) discuss the scalability of our approach.

\subsection{Text Prompt}

Here we give the text prompts of the three foreground images shown in Fig.~5 of the comparative experiment to facilitate the reproduction of the results. They are described by the most advanced AIGC model, GPT-4, as follows:

\begin{itemize}
    \item A bowl of stir-fried golden-brown cauliflower garnished with red and green chili slices, served in a patterned black and white bowl. The dish is lightly coated with spice, presenting a simple yet visually appealing and spicy flavor, high quality, 4k.
    \item Two golden-brown, sesame-coated pastries served on a small, brightly colored plate with an orange-red rim, high quality, 4k.
    \item A bowl of steaming hot noodles topped with tender beef chunks, green vegetables,  diced carrots in a rich broth, served in a white bowl, high quality, 4k.
\end{itemize}

\subsection{More Results}

% Fig.~\ref{fig:more_result} presents additional visual results generated by our proposed method, Foodfusion. As illustrated in the figure, our approach successfully achieves high-quality and realistic image composition across various foreground and background combinations, even without explicit location information, such as masks.

Fig.~\ref{fig:more_result} showcases additional visual results produced by our proposed method, Foodfusion, further highlighting its exceptional performance in food image synthesis. The results vividly demonstrate the method's ability to generate high-quality, realistic composite images across diverse foreground and background pairings, effectively capturing intricate details and natural aesthetics. Notably, this high level of synthesis quality is achieved even without explicit spatial guidance, such as masks, underscoring the robustness and adaptability of our approach. This capability allows Foodfusion to handle complex food image composition scenarios with remarkable precision, maintaining consistency and coherence in the synthesized images. This is critical for applications such as digital advertising, food photography, and other visually demanding contexts. The method’s inherent ability to generalize across varying conditions and produce seamless compositions further reinforces its potential as a powerful tool in image composition.

\subsection{Expansion Discussion}

In the main paper, we demonstrate the superior performance of our proposed method, Foodfusion, in the specific context of food image composition. To further validate the generalizability of Foodfusion to other image composition tasks, we provide additional experimental results in this subsection.

Fig.~\ref{fig:people} illustrates the performance of our method across various image composition scenarios, using familiar real-world portrait images as examples. The results demonstrate that our proposed method can effectively composite portrait images, highlighting its strong generalization capabilities and excellent performance in different image composition tasks.

% \end{document}

\end{document}